\documentclass[sigconf]{acmart}

\usepackage{multirow}
\usepackage{colortbl}
\usepackage{multirow}
\usepackage{algorithm}
\usepackage{algorithmicx}
\usepackage{algpseudocode}

\floatname{algorithm}{Algorithm}

\usepackage[normalem]{ulem}
\useunder{\uline}{\ul}{}
\AtBeginDocument{%
  }

\copyrightyear{2025} 
\acmYear{2025} 
\setcopyright{cc}
\setcctype{by}
\acmConference[MM '25]{Proceedings of the 33rd ACM International Conference
 on Multimedia}{October 27--31, 2025}{Dublin, Ireland}
\acmBooktitle{Proceedings of the 33rd ACM International Conference on
 Multimedia (MM '25), October 27--31, 2025, Dublin,
 Ireland}
\acmDOI{10.1145/3746027.3755275}
\acmISBN{979-8-4007-2035-2/2025/10}

\begin{document}

\title{T2VParser: Adaptive Decomposition Tokens for Partial Alignment in Text to Video Retrieval}

\author{Yili Li}
\affiliation{%
  \institution{Institute of Information Engineering, Chinese Academy of Sciences}
  \institution{School of Cyber Security, University of Chinese Academy of Sciences}
  \city{Beijing}
  \country{China}}
\email{liyili@iie.ac.cn}
\orcid{0009-0006-4037-819X}

\author{Gang Xiong}
\affiliation{%
  \institution{Institute of Information Engineering, Chinese Academy of Sciences}
  \city{Beijing}
  \country{China}}
\email{xionggang@iie.ac.cn}
\orcid{0000-0002-3190-6521}

\author{Gaopeng Gou}
\authornote{Corresponding authors}
\affiliation{%
  \institution{Institute of Information Engineering, Chinese Academy of Sciences}
  \city{Beijing}
  \country{China}}
\email{gougaopeng@iie.ac.cn}
\orcid{0000-0002-3533-4874}

\author{Xiangyan Qu}
\affiliation{%
  \institution{Institute of Information Engineering, Chinese Academy of Sciences}
  \city{Beijing}
  \country{China}}
\email{quxiangyan@iie.ac.cn}
\orcid{0000-0003-3658-8099}

\author{Jiamin Zhuang}
\affiliation{%
  \institution{Institute of Information Engineering, Chinese Academy of Sciences}
  \city{Beijing}
  \country{China}}
\email{zhuangjiamin@iie.ac.cn}
\orcid{0000-0002-1598-0764}

\author{Zhen Li}
\affiliation{%
  \institution{Institute of Information Engineering, Chinese Academy of Sciences}
  \city{Beijing}
  \country{China}}
\email{lizhen@iie.ac.cn}
\orcid{0000-0002-3892-4909}

\author{Junzheng Shi}
\affiliation{%
  \institution{Institute of Information Engineering, Chinese Academy of Sciences}
  \city{Beijing}
  \country{China}}
\email{shijunzheng@iie.ac.cn}
\orcid{0000-0003-4653-1686}

\renewcommand{\shortauthors}{Yili Li et al.}

\begin{abstract}

Text-to-video retrieval essentially aims to train models to align visual content with textual descriptions accurately.  Due to the impressive general multimodal knowledge demonstrated by image-text pretrained models such as CLIP, existing work has primarily focused on extending CLIP knowledge for video-text tasks. However, videos typically contain richer information than images. In current video-text datasets, textual descriptions can only reflect a portion of the video content, leading to partial misalignment in video-text matching. Therefore, directly aligning text representations with video representations can result in incorrect supervision, ignoring the inequivalence of information. In this work, we propose T2VParser to extract multiview semantic representations from text and video, achieving adaptive semantic alignment rather than aligning the entire representation. To extract corresponding representations from different modalities, we introduce Adaptive Decomposition Tokens, which consist of a set of learnable tokens shared across modalities. The goal of T2VParser is to emphasize precise alignment between text and video while retaining the knowledge of pretrained models. Experimental results demonstrate that T2VParser achieves accurate partial alignment through effective cross-modal content decomposition. The code is available at \href{https://github.com/Lilidamowang/T2VParser}{https://github.com/Lilidamowang/T2VParser}.

\end{abstract}

\maketitle

\section{Introduction}
\label{sec:intro}
Large-scale visual-language pre-training has significantly advanced cross-modal representation learning through training on extensive multimodal data \cite{clip} \cite{frozen_in_time} \cite{unpaired_data_for_vltraining}. For video-text retrieval, recent efforts have focused on transferring knowledge from image-text models due to limited video-text data and high computational costs, achieving state-of-the-art results \cite{clip-vip} \cite{STAN}. CLIP4Clip \cite{clip4clip} pioneered this approach using mean pooling of CLIP frame representations \cite{clip}. Subsequent works like STAN \cite{STAN} and CLIP-VIP \cite{clip-vip} enhanced video representation by incorporating temporal modeling modules into CLIP's visual encoder. From a comprehensive perspective, existing methods mainly extend the image-text alignment, focusing on minimizing the distance between video and textual features.

\begin{figure}[t]
  \centering
  \includegraphics[width=\linewidth]{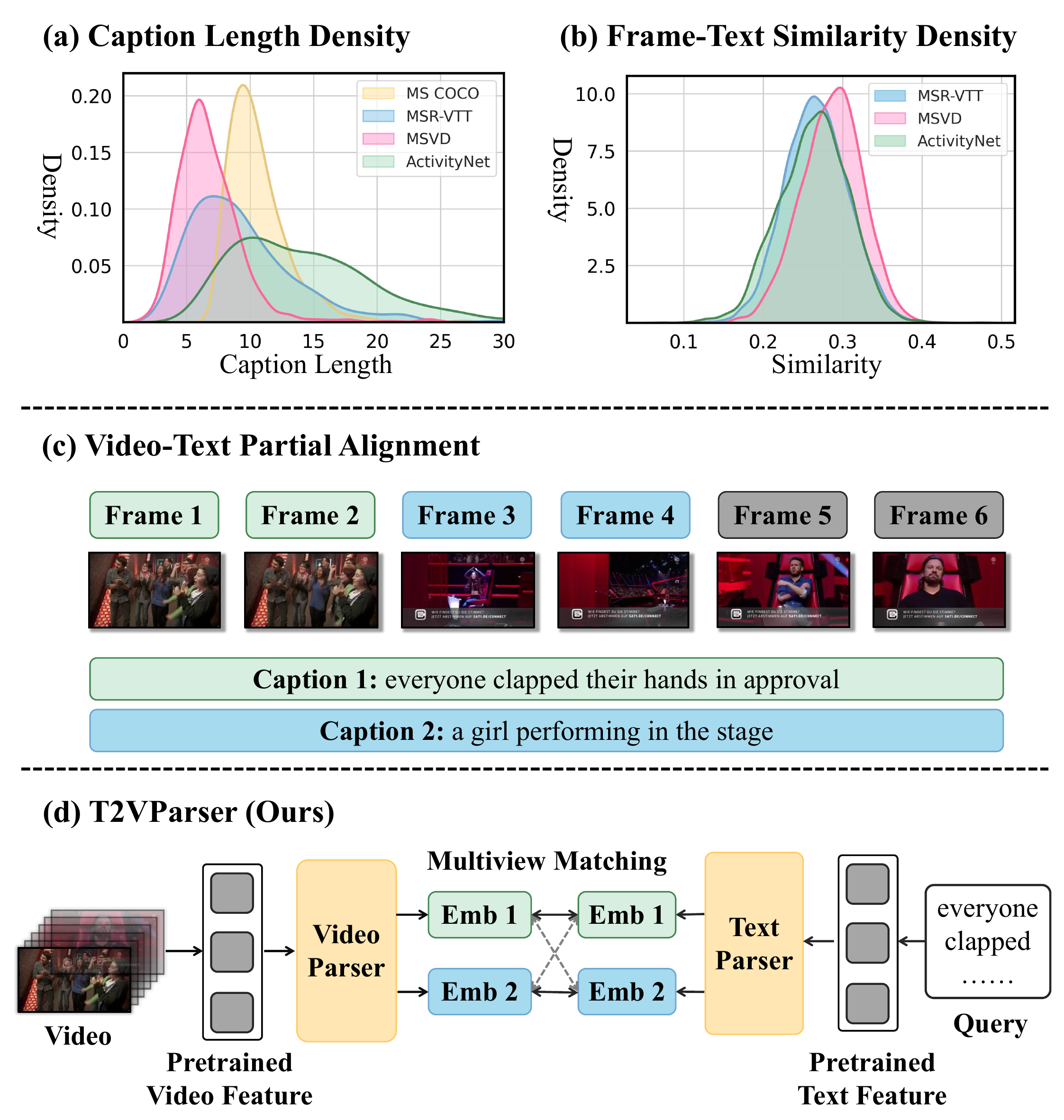}
  \caption{(a) Text length distributions; (b) Frame-text similarity; (c) Partial alignment example; (d) T2VParser framework.}
  \label{fig:mov}
  \vspace{-10pt}
\end{figure}

Videos inherently encapsulate richer semantic content and complex temporal dynamics compared to static images. However, as illustrated in Figure \ref{fig:mov}(a), video descriptions in benchmarks are frequently shorter than or comparable to image captions, resulting in significant information gaps. This limitation leads to partial alignment scenarios, where textual descriptions correspond exclusively to specific frames while neglecting others, as demonstrated in Figure \ref{fig:mov}(c). 
Quantitative analysis employing CLIP's text-frame similarity metrics, as illustrated in Figure \ref{fig:mov}(b), reveals that approximately half of the video frames exhibit lower similarity scores with their corresponding descriptions compared to the dataset averages (0.26 in MSR-VTT, 0.28 in MSVD, and 0.26 in ActivityNet). This phenomenon demonstrates the information inequivalence between videos and their descriptions, indicating that a video may correspond to diverse descriptions and contents. So, it is necessary to use semantically different embeddings for partial alignment.

Several methods have been proposed to precise video-text alignment. Early works like \cite{HGR} \cite{HANet} used explicit partial alignment (frame-phrase matching) but lacked semantic depth. \cite{VPM} improved semantics via implicit multi-frame aggregation, though disrupting encoder integrity.
With the image-text pre-training has demonstrated significant advantages, X-Pool \cite{x-pool} uses attention mechanisms to focus on specific frames, achieving alignment at the cost of visual information loss. Meanwhile, token-wise matching methods using pretrained encoder like Mug-STAN \cite{mug-stan} and CLIP-VIP \cite{clip-vip} aligned text tokens with individual frames. But they consider each token as an independent semantic unit and overlooked temporal coherence and global video understanding.

We aim for precise text-video alignment while preserving pretrained model knowledge via representation decomposition. We propose \textbf{A}daptive \textbf{D}ecomposition \textbf{T}okens (ADTs) as distinct queries to adaptively retrieve textual and video information into semantically corresponding multiview embeddings. This framework is named \textbf{T2VParser}. T2VParser uses modality-shared learnable tokens to extract multiperspective semantic content, generating embeddings where same token outputs align crossmodality (Figure \ref{fig:mov} (d)). The Dual Communication Mechanism enables embedding exchange/filtering to reduce interference and enhance alignment. Doc-Video training with diversity loss ensures representation diversity. Compared to explicit and implicit alignment, our method preserves the semantic understanding capability of the pretrained encoder. Distinct from token-wise interaction, we introduce temporally coherent local representations (multiview embeddings).


In conclusion, our contributions are threefold: 
(1) We studied the differences between video-text and image-text data and analyzed the issue of partial alignment. To enhance video-text alignment while preserving pretrained encoder capabilities, learnable modality-shared Adaptive Decomposition Tokens (ADTs) were proposed, enabling decomposition of text and video into semantically independent multiview embeddings.
(2) A Dual Communication Mechanism was developed to facilitate information exchange between independent representations, filtering out semantically irrelevant elements and focusing on alignment-relevant features.
(3) T2VParser is model-agnostic and universal to cooperatewith various independent encoder approaches. Experiments demonstrate T2VParser achieve state-of-the-art performance with different encoder and data formats, and the advantage becoming increasingly pronounced when processing more detailed and rich data.
\vspace{-10pt}

\section{Related Work}
\label{sec:related}
\noindent \textbf{Text-video retrieval.}
Current video-text retrieval methods focus on creating a joint embedding space where similar video and text vectors are closely positioned for retrieval. The field has progressed through two stages: 1) Explicit alignment (e.g., HGR \cite{HGR}, HANet \cite{HANet}) matching global/local features via expert tools like object detection; 2) Image knowledge transfer using models like CLIP \cite{clip}. CLIP4Clip \cite{clip4clip} averaged CLIP frame embeddings, while STAN \cite{STAN} and CLIP-VIP \cite{clip-vip} added temporal modeling. However, these methods assume full video-text correspondence, causing partial alignment errors. Advanced pretrained representations and novel mechanisms are needed for robust alignment.
\noindent \textbf{Partial alignment in text-video data.}
Partial alignment arises when text reflecting only partial information is matched with entire videos, potentially generating incorrect supervision signals that hinder model learning. While not explicitly designed for this issue, local alignment methods like \cite{HGR} and \cite{HANet} mitigate its impact through frame-level rather than global video-text matching.
Several studies have tackled partial alignment through dataset and model innovations. For instance, MQVR \cite{MQVR} introduces a many-to-many evaluation metric to address fairness but cannot prevent incorrect alignment learning. \cite{VPM} extracts local embeddings from video frames for partial alignment, though it compromises pretrained encoder capabilities. X-Pool \cite{x-pool} uses attention mechanisms to focus on specific frames, achieving alignment at the cost of visual information loss. Similarly, Mug-STAN \cite{mug-stan} and CLIP-VIP \cite{clip-vip} implement token-wise matching, treating each token as an independent unit but ignoring information coherence.
Effective video-text alignment thus requires leveraging pretrained vision-text encoders while addressing semantic discrepancies caused by partial alignment.
\begin{figure*}[ht]
  \centering
  \includegraphics[width=\linewidth]{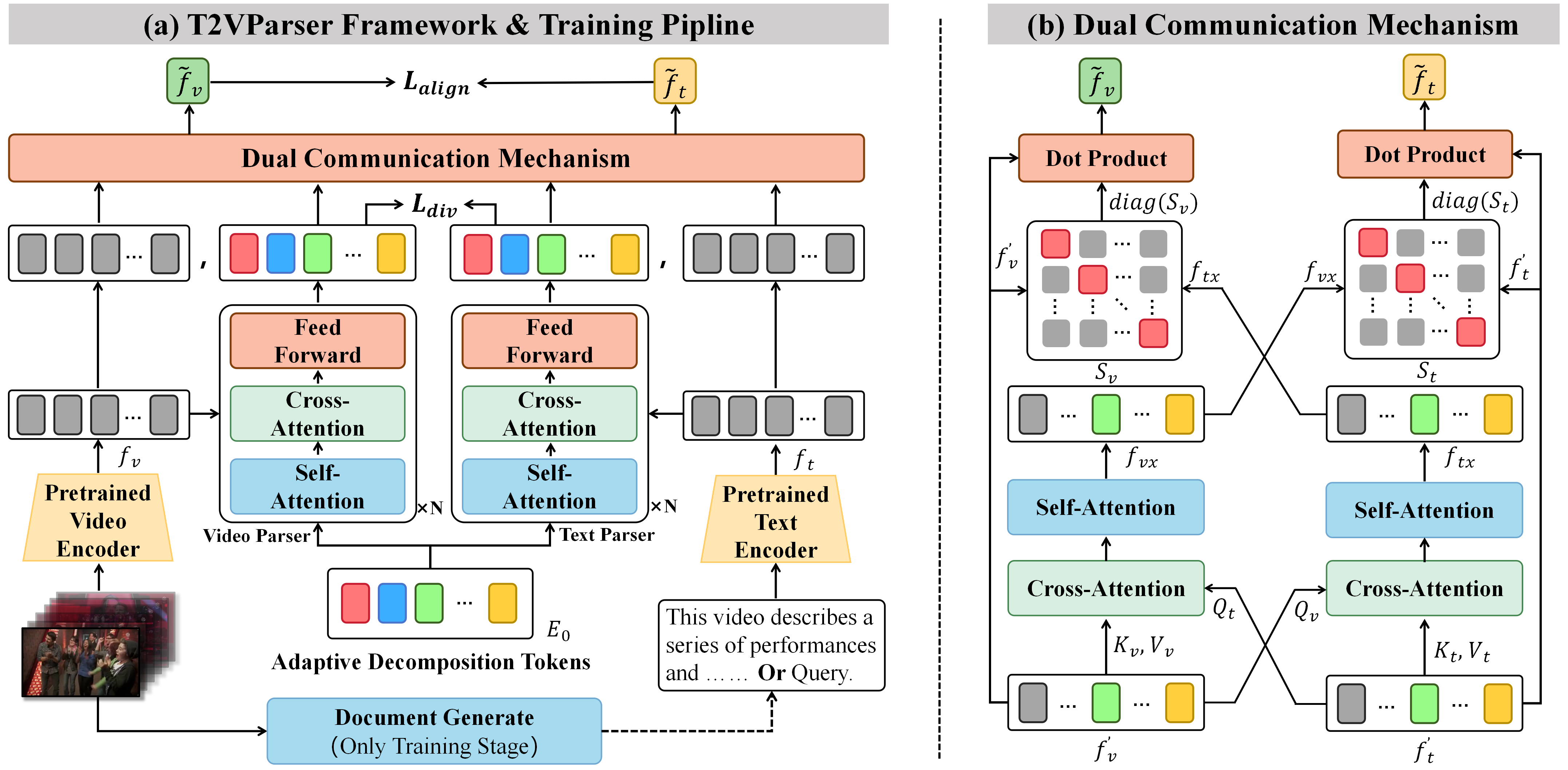}
  \caption{An overview of our T2VParser. (a) illustrates the processing and training pipeline of T2VParser. It shows how the Video Parser and Text Parser obtain multiview embeddings for different modality data, which are then aligned through the Dual Communication Mechanism by acquiring supplementary information between them. (b) shows the details of Dual Communication Mechanism. This module aims to facilitate the interaction of different modality and comprehensively measure the similarity between video and text multiview embeddings.}
  \label{fig:model}
  \vspace{-10pt}
\end{figure*}

\section{Methodology}
\label{sec:methodology}

Text-to-video retrieval evaluates multimodal alignment by retrieving videos $\{ v \} \in V$ relevant to query $t$ via similarity scoring. While conventional methods align modality representations using paired data, our T2VParser framework enhances partial alignment through adaptive semantic decomposition. As shown in Figure \ref{fig:model}, it decomposes video $f_v$ and text $f_t$ into multiview embeddings ($f_v^{'}$, $f_t^{'}$) using Adaptive Decomposition Tokens (ADTs) that preserve pretrained encoder functionality. The Dual Communication Mechanism optimizes cross-modal information flow, while Doc-Video training with enriched texts addresses semantic sparsity. Multiview modality alignment was achieved through joint optimization of contrastive learning and representation diversity loss functions.

\subsection{Adaptive Decomposition Tokens}
The primary objective of this work is to achieve precise multimodal alignment through semantic decomposition across modalities. While existing approaches have successfully incorporated temporal information into video representations to establish coarse multimodal alignment \cite{STAN}, our methodology focuses on the controlled decomposition of these representations into distinct perspectives, and then achieve partial alignment based on semantic relationships between them. Initially, text and video representations are extracted using pretrained encoders. Subsequently, drawing inspiration from DETR \cite{DRET} in object detection, modality-shared learnable tokens are employed as event queries from multiple perspectives. These tokens iteratively retrieve relevant content from both video and text representations, ensuring encoder effectiveness while enabling adaptive perspective parsing from diverse content.

\noindent \textbf{Video and Text Representations.} 
To ensure the extraction of high-quality multiview embeddings, the original representations must encompass rich and accurate multimodal information while effectively capturing temporal dependencies. The pretrained STAN encoder \cite{STAN} \cite{mug-stan} was selected as the feature extractor to meet these requirements. Given a video sequence $v_i =\{ v_i^1, v_i^2, ..., v_i^N \}$ (where $N$ denotes frame count) and a text query $t_i = \{ t_i^1, t_i^2, ..., t_i^M \}$ (where $M$ represents token count), the corresponding representations $f_v = \{f_v^{CLS}, f_v^1, f_v^2, ..., f_v^N\}$ and $f_t = \{ f_t^{CLS}, f_t^1, f_t^2, ... , f_t^M \}$ are generated. Each frame representation is enriched with contextual content, ensuring comprehensive temporal modeling.

\noindent \textbf{Content Parser with Adaptive Decomposition Tokens.} 
As depicted in Figure \ref{fig:mov}, video and text semantics are derived from multiple perspectives, with only partial perspective alignment occurring during matching. To integrate multimodal representations and extract multiview embeddings while preserving pretrained encoder, a Video Parser and Text Parser were developed, interconnected through modality-shared Adaptive Decomposition Tokens.

As shown in Figure \ref{fig:model} (a), to achieve adaptive and controlled acquisition of representations from different perspectives, we introduce a set of modality-shared learnable tokens $E_0=\{e_0^1, e_0^2, ..., e_0^k\}$ as inputs to the Parsers, where $k$ is the number of learnable tokens. Each $e_0$ can be considered as a distinct query that retrieves corresponding information from the representations. Taking the Video Parser as an example, the text-side follows a similar process. Initially, $E_0$ is input and processed through self-attention to obtain $\hat{E}_0$, facilitating information interaction among each $e_0$,
\begin{equation}
    \hat{E}_0=\text{Self-Attention}(e_0^1, e_0^2, ..., e_0^k;\theta_{1}^{sa})
    \label{eq:sa}
\end{equation}
where $\theta_{1}^{sa}$ represents the parameters of self-attention in the first video parser layer. Subsequently, $\hat{E}_0$ serves as the query, and through a cross-attention mechanism, semantic information corresponding to each perspective of $\hat{e}_0$ is retrieved from the video representations $f_v = \{f_v^{CLS}, f_v^1, f_v^2, ..., f_v^N\}$. In this process, the video representations $f_v$ are transformed into $K_0=Linear(f_v; \theta_0^{k})$ and $V_0=Linear(f_v; \theta_0^{v})$ using parameters $\theta_0^{k}$ and $\theta_0^{v}$, respectively. Similarly, $\hat{E}_0$ is transformed into $Q_0=Linear(\hat{E}_0; \theta_0^{q})$ using parameter $\theta_0^{q}$. The process of multiview representation retrieval is as follows:
\begin{equation}
    E'_{0} = softmax(\frac{Q_0 K_0^T}{\sqrt{d_k}}) V_0
    \label{eq:ca}
\end{equation}
where $d_k$ denotes the embedding dimension. $E'_0$ is subsequently processed through a fully connected layer, followed by layer normalization and GELU, generating the first layer output $E_1$. A residual connection is established by adding $E_1$ to the previous output,
\begin{equation}
    E_{1} = Linear(E'_0) + E'_0
    \label{eq:ff}
\end{equation}

For the N-layer Video Parser, the flow is same as the first layer. After completing the final layer of the Video Parser, we obtain $E_N$. We then add it to the normalized global representation of $f_v$, denoted as $f_v^{CLS}$, and subsequently concatenate it with other local-level representations $f_v^{local} = \{f_v^1, f_v^2, ..., f_v^N\}$ to obtain the final multiview representation on the video side, $f'_v \in \mathbb{R}^{(k+N) \times d}$,
\begin{equation}
    E_N^{CLS} = norm(E_N) + norm(f_v^{CLS})
    \label{eq:mv}
\end{equation}
\begin{equation}
    f'_v = concat(f_v^{local}, E_N^{CLS})
\end{equation}
The final representations $f'_v$ and $f'_t$ integrate global, token/frame-level, and perspective-specific information, such as events, actions, and subjects. These foundational embeddings serve as the basis for subsequent semantic matching.

\subsection{Dual Communication Mechanism}
Target video retrieval requires alignment and similarity computation between $f'_v$ and $f'_t$ representations. While direct global alignment with averaging or individual similarity computation followed by score averaging (as in \cite{HGR}) represent potential approaches, these methods are susceptible to inaccuracies from partial alignment, as demonstrated in Figure \ref{fig:mov} (c). This section investigates the filtering of retrieval-relevant information from multiview embeddings to achieve effective representation-level partial alignment.

We propose the Dual Communication Mechanism. Using video multiview embeddings $f'_v$ as an example (with symmetric text-side processing), cross-attention is first computed to extract supplementary information from text modality $f'_t$. This information is then integrated through self-attention, producing $f_{vx}$,
\begin{equation}
    S = f'_v \cdot (f'_t)^T
\end{equation}
\begin{equation}
    S_{t2v} = softmax(S)
\end{equation}
\begin{equation}
    f_{vx} = \text{Self-Attention}(S_{t2v} \cdot f'_v)
\end{equation}
where $f_{vx}$ represents the video cross-modal representations obtained from $f'_v$, each corresponding to a text embedding based on the semantics of $f'_t$. Similarly, $f_{tx}$ follows the same principle.

To identify alignment-relevant information for retention, a similarity matrix between $f'_v$ and $f_{tx}$ is computed, with diagonal elements extracted as similarity scores $S_v$, representing correspondence between video multiview embeddings and video-content-fused text multiview embeddings.
\begin{equation}
    S_{v} = softmax(diag(f'_v \cdot f_{tx}))
\end{equation}

In the end, we sum $f'_v$ based on the weights in $S_v$ to obtain $\widetilde{f}_v$,
\begin{equation}
    \widetilde{f}_v = S_{v} \cdot f'_v
\end{equation}
$\widetilde{f}_t$ is similarly computed. These final representations integrate multiview information while maintaining precise partial alignment.

\subsection{Document Generation}
\label{docgen}

\begin{figure}[t]
  \centering
  \includegraphics[width=\linewidth]{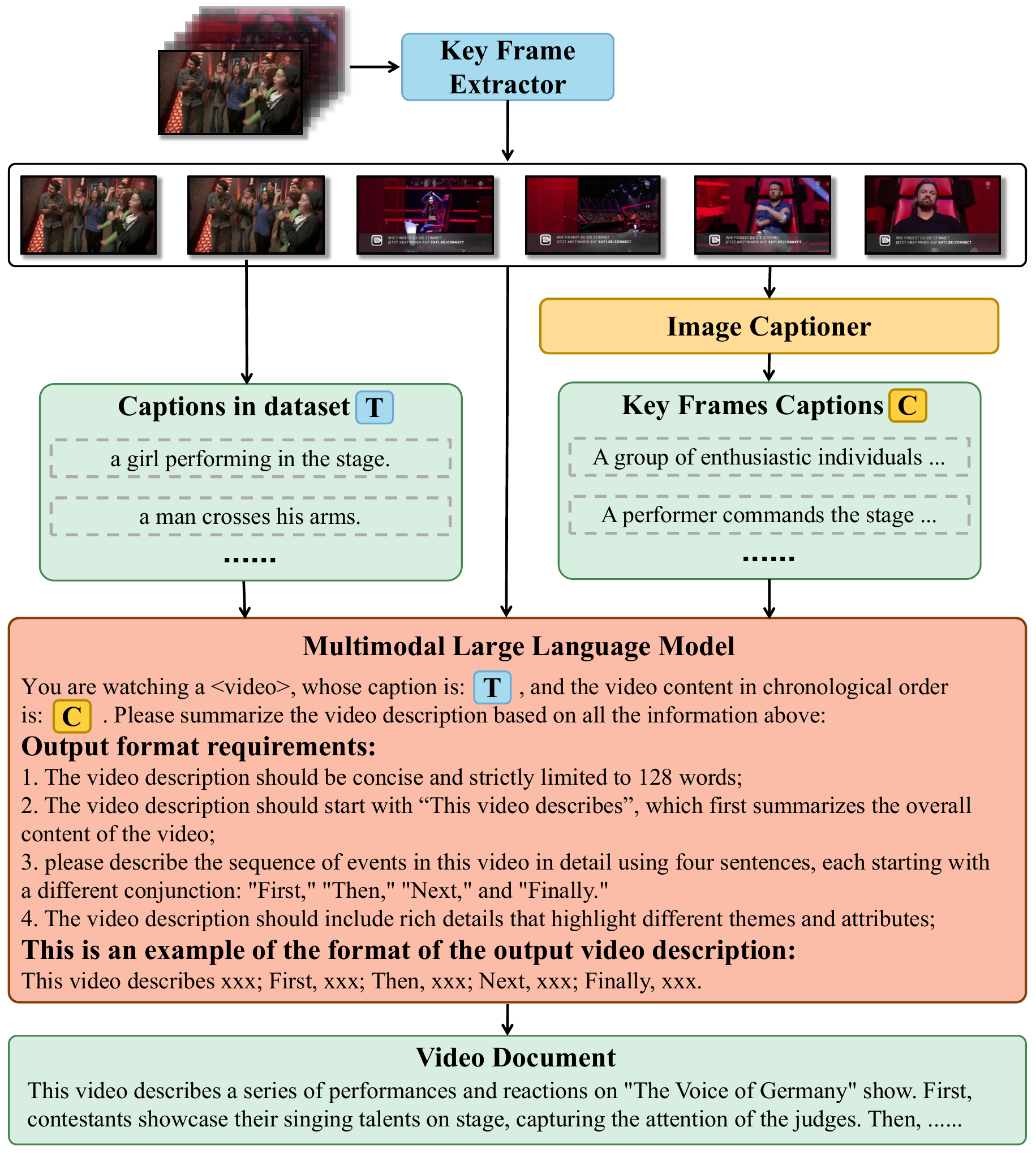}
  \caption{The pipeline of video document generation.}
  \label{fig:doc}
  \vspace{-10pt}
\end{figure}

Accurate multiview embedding extraction requires information-rich original representations. While the Video Parser decomposes necessary information from videos, current text-video datasets lack sufficient text length for effective decomposition (Figure \ref{fig:mov} (a)). To enhance multiview embeddings, we provide semantically rich video documents as additional training data (Figure \ref{fig:doc}).

\noindent \textbf{Key-Frames Extraction and Captioning.} 
Videos contain rich but sparse information, with many irrelevant frames. To create accurate video documents, we first extract 8 key frames per video using TSDPC (Temporal Segment Density Peaks Clustering) \cite{TSDPC}, then generate descriptions for each frame with BLIP-2 \cite{BLIP-2}.

\noindent \textbf{Document Generation via LLM.}
Documents $D$ with comprehensive semantic information are generated using provided descriptions $T =\{ t_1, t_2,... \}$ and key frame descriptions $C=\{c_1, c_2,..., c_8 \}$ as input to Deepseek-v2 \cite{deepseek} Large Language Model. The prompt construction and final input format are illustrated in Figure \ref{fig:doc}.

\subsection{Training Strategy and Constraints}

\noindent \textbf{Doc-Video Training.}
Based on the text descriptions $T=\{t_1, t_2,... \}$ provided in the dataset and the generated video documents $D=\{d_1, d_2, ...\}$, we can construct a query set $Q_i=\{ T_i, D_i \}$ for each video $v_i$, where both $T_i$ and $D_i$ represent the description sets corresponding to video $v_i$. We can represent the training data pairs for the T2VParser as $\{ q,v_i \}$, where $q \in Q_i$ and $v_i \in V$.

\noindent \textbf{Representation Alignment Loss.}
Given $\widetilde{f}_t$ and $\widetilde{f}_v$, we compute their similarity and optimize alignment performance within a batch,
\begin{equation}
    \mathcal{L}_{v2t}=-\frac{1}{B}\sum_{i=1}^{B}\log\frac{\exp(s(\widetilde{f_v^{i}}, \widetilde{f}_t^i))}{\sum_{j=1}^{B}\exp(s(\widetilde{f_v^{i}}, \widetilde{f}_t^j))}
\end{equation}
\begin{equation}
    \mathcal{L}_{t2v}=-\frac{1}{B}\sum_{i=1}^{B}\log\frac{\exp(s(\widetilde{f_v^{i}}, \widetilde{f}_t^i))}{\sum_{j=1}^{B}\exp(s(\widetilde{f_v^{j}}, \widetilde{f}_t^i))}
\end{equation}
\begin{equation}
    \mathcal{L}_{align} = \mathcal{L}_{v2t} + \mathcal{L}_{t2v}
\end{equation}
For the extracted representations $\widetilde{f}_t$ and $\widetilde{f}_v$ relevant to the retrieval, the constraint $\mathcal{L}_{align}$ can make their representations more similar, allowing for direct similarity calculation and ranking using $\widetilde{f}_t$ and $\widetilde{f}_v$ during the inference phase. Additionally, for the multiview embeddings $f'_v \in \mathbb{R}^{(k+N) \times d}$ and $f'_t \in \mathbb{R}^{(k+M) \times d}$, $\mathcal{L}_{align}$ can also achieve partial alignment through $\widetilde{f}_t$ and $\widetilde{f}_v$.

\noindent \textbf{Representation Diversity Loss.}
Experiments reveal that using only alignment loss may cause feature collapse \cite{VSD}, reducing embedding diversity and impairing the Dual Communication Mechanism's ability to integrate multi-perspective information. To preserve distinct viewpoints, we introduce a representation diversity loss to maximize inter-embedding variance. Both Video and Text Parsers follow this approach.
For $E_N \in \mathbb{R}^{k \times d}$, we first compute their similarity matrices with themselves respectively,
\begin{equation}
    M=E_N \cdot (E_N)^T
\end{equation}
\begin{equation}
    \mathbb{I}=diag(M), \in \mathbb{R}^{k \times k}
\end{equation}
\begin{equation}
    M'=M-\mathbb{I}
\end{equation}
and we get $M'_t$ and $M'_v$. The final $\mathcal{L}_{div}$ is the sum of the similarities within each set of multiview embeddings, with lower values indicating lower similarity with others and greater diversity.
\begin{equation}
    \mathcal{L}_{div}=\frac{1}{2}(||M'_t||^2 + ||M'_v||^2)
\end{equation}

\section{Experiments}
\label{sec:experiments}
\noindent \textbf{Datasets and Evaluation Metrics}
We evaluated our model on four benchmarks: (1) MSR-VTT \cite{msr-vtt}: 10K videos (200K captions), tested on Training-9k split \cite{Multi-modal-Transformer}. (2) MSVD \cite{msvd}: 1,970 videos (around 40 captions each), divided into 1,200 training, and 670 test videos. (3) DiDeMo \cite{didemo}: 10K videos (40K sentences), using paragraph retrieval setting \cite{use_what_you_have} \cite{less_is_more} \cite{frozen_in_time}. (4) ActivityNet \cite{activitynet}: 20K videos with paragraph retrieval \cite{Cross_modal_and_hierarchical_modeling}.
We adopt standard retrieval metrics, namely recall at rank K (R@K), calculates the percentage of instances where the correct result is successfully retrieved within top K.

\noindent \textbf{Implementation Details.}
For each video, we sample 12 frames (64 for long videos) at 224×224 resolution. Text tokens are fixed at 128 length. Both Video and Text Parsers use 8 layers with residual connections and 8 ADTs. The final loss function is a weighted sum, $\mathcal{L} = \mathcal{L}_{align} + \alpha \mathcal{L}_{div}$, where $\alpha$ is set to 0.1. Training uses $Lr=10^{-6}$ (encoders) and $10^{-5}$ (others). Besides, the extra captions only use in training stage, and we also set some ablation studies about Doc-Video Training in Sec \ref{ablms} for fairness.

\begin{table*}[ht]
\caption{Evaluation of Long-Text Retrieval Performance. We report the R@1, R@5, and R@10 metrics for the text-to-video task on various datasets. {\color[HTML]{000AFF} Blue} indicates the average length of the text in each dataset. MSR-VTT-Doc and MSVD-Concat are self-created long-text retrieval datasets.}
\label{tab:sd}
\vspace{-5pt}
\resizebox{\linewidth}{!}{
\begin{tabular}{lcccccccccccc}
\hline
\multicolumn{1}{l|}{\multirow{2}{*}{\textbf{Method}}} & \multicolumn{3}{c|}{\textbf{MSR-VTT-1k({\color[HTML]{000AFF} 9.3})}}                       & \multicolumn{3}{c|}{\textbf{DiDeMo({\color[HTML]{000AFF} 29.5})}}                          & \multicolumn{3}{c|}{\textbf{MSR-VTT-Doc({\color[HTML]{000AFF} 107.1})}}                    & \multicolumn{3}{c}{\textbf{MSVD-Concat({\color[HTML]{000AFF} 288.3})}} \\ \cline{2-13} 
\multicolumn{1}{l|}{}                                 & \textbf{R@1↑} & \textbf{R@5↑} & \multicolumn{1}{c|}{\textbf{R@10↑}} & \textbf{R@1↑} & \textbf{R@5↑} & \multicolumn{1}{c|}{\textbf{R@10↑}} & \textbf{R@1↑} & \textbf{R@5↑} & \multicolumn{1}{c|}{\textbf{R@10↑}} & \textbf{R@1↑}  & \textbf{R@5↑} & \textbf{R@10↑} \\ \hline
\multicolumn{13}{c}{\textbf{Global Matching Baseline}}                                                                                                                                                                                                                                                                    \\ \hline
\multicolumn{1}{l|}{CLIP4Clip \cite{clip4clip}}                        & 44.5          & 71.0          & \multicolumn{1}{c|}{81.6}           & 43.4          & 70.1          & \multicolumn{1}{c|}{80.1}           & 76.5          & 88.2          & \multicolumn{1}{c|}{90.4}           & 81.9           & 91.3          & 95.2           \\
\multicolumn{1}{l|}{T2VParser+CLIP4Clip}              & 46.4          & 73.6          & \multicolumn{1}{c|}{82.4}           & 45.4          & 72.4          & \multicolumn{1}{c|}{83.7}           & 81.2          & 90.9          & \multicolumn{1}{c|}{91.3}           & 86.9           & 93.6          & 99.7           \\
\multicolumn{1}{l|}{Improvement}                      & \textbf{+2.4} &               & \multicolumn{1}{c|}{}               & \textbf{+2.0} &               & \multicolumn{1}{c|}{}               & \textbf{+4.7} &               & \multicolumn{1}{c|}{}               & \textbf{+5.0}  &               &                \\ \hline
\multicolumn{1}{l|}{STAN \cite{STAN}}                             & 54.1          & 79.5          & \multicolumn{1}{c|}{87.8}           & 54.6          & 78.4          & \multicolumn{1}{c|}{85.1}           & 84.3          & 90.6          & \multicolumn{1}{c|}{94.7}           & 86.1           & 93.3          & 95.5           \\
\multicolumn{1}{l|}{T2VParser+STAN}                   & 55.8          & 79.9          & \multicolumn{1}{c|}{88.3}           & 56.4          & 81.7          & \multicolumn{1}{c|}{87.2}           & 87.8          & 92.4          & \multicolumn{1}{c|}{95.7}           & 89.9           & 97.2          & 99.7           \\
\multicolumn{1}{l|}{Improvement}                      & \textbf{+1.7} &               & \multicolumn{1}{c|}{}               & \textbf{+1.8} &               & \multicolumn{1}{c|}{}               & \textbf{+3.5} &               & \multicolumn{1}{c|}{}               & \textbf{+3.8}  &               &                \\ \hline
\multicolumn{13}{c}{\textbf{Token-wise Matching Baseline}}                                                                                                                                                                                                                                                                \\ \hline
\multicolumn{1}{l|}{Mug-STAN \cite{mug-stan}}                         & 57.3          & 81.6          & \multicolumn{1}{c|}{88.4}           & 61.2          & 84.1          & \multicolumn{1}{c|}{88.9}           & 90.2          & 99.1          & \multicolumn{1}{c|}{99.9}           & 91.3           & 99.4          & 99.7           \\
\multicolumn{1}{l|}{T2VParser+Mug-STAN}               & 58.4          & 82.1          & \multicolumn{1}{c|}{88.4}           & 62.4          & 85.2          & \multicolumn{1}{c|}{89.5}           & 92.9          & 99.3          & \multicolumn{1}{c|}{99.9}           & 94.6           & 99.4          & 100.0          \\
\multicolumn{1}{l|}{Improvement}                      & \textbf{+1.1} &               & \multicolumn{1}{c|}{}               & \textbf{+1.2} &               & \multicolumn{1}{c|}{}               & \textbf{+2.7} &               & \multicolumn{1}{c|}{}               & \textbf{+3.3}  &               &                \\ \hline
\multicolumn{1}{l|}{CLIP-VIP \cite{clip-vip}}                         & 57.7          & 80.5          & \multicolumn{1}{c|}{88.2}           & 55.3          & 82.0          & \multicolumn{1}{c|}{89.3}           & 90.2          & 98.7          & \multicolumn{1}{c|}{99.9}           & 91.8           & 99.1          & 100.0          \\
\multicolumn{1}{l|}{T2VParser+CLIP-VIP}               & 58.4          & 82.1          & \multicolumn{1}{c|}{88.1}           & 57.0          & 85.1          & \multicolumn{1}{c|}{89.6}           & 92.9          & 99.1          & \multicolumn{1}{c|}{99.9}           & 94.8           & 99.4          & 100.0          \\
\multicolumn{1}{l|}{Improvement}                      & \textbf{+0.7}  &               & \multicolumn{1}{c|}{}               & \textbf{+1.7} &               & \multicolumn{1}{c|}{}               & \textbf{+2.7} &               & \multicolumn{1}{c|}{}               & \textbf{+3.0}  &               &                \\ \hline
\end{tabular}
}
\end{table*}

\subsection{Advantage in Semantic Decomposition}
\label{4.1}
T2VParser effectively disentangles intertwined information from multiple levels (token-level and event-level) and achieves partial semantic alignment. To systematically evaluate its capability in processing complex descriptions and video content, we constructed a long-text video retrieval test in Table \ref{tab:sd}. Specifically, for MSR-VTT-Doc, we generated 128-token description documents for each video in MSR-VTT following the methodology in Section \ref{docgen}, while maintaining the original split for training and testing. For DiDeMo, the official annotations provide multiple detailed descriptions in chronological order, which can be combined into information-rich documents for training and testing, as in the standard testing methods used by \cite{use_what_you_have}, \cite{less_is_more}, and \cite{frozen_in_time}. Therefore, we evaluate following their approach, where all sentence descriptions for a video are concatenated into a document. For MSVD-Concat, we similarly combined the descriptions of each MSVD video. 

We selected two types of baseline models for evaluation: Global Matching and Token-wise Matching. As shown in Table \ref{tab:sd}, T2VParser improves the baselines under different encoder architectures and data formats. Notably, the performance gain increases as the data complexity grows. Under the text-scarce MSR-VTT-1k setting, the improvement is around 1.0\%, while with more comprehensive textual information, the gain increases to 3. 0\%. Further analysis reveals that this improvement stems from the Adaptive Decomposition Tokens effectively disentangling information. With increasing text complexity, Adaptive Decomposition Tokens enable precise retrieval through multiview information extraction and partial alignment. While token-wise matching models perform adequately with limited text, their neglect of event-level integrity creates semantic gaps, resulting in more performance gap with T2VParser.

\subsection{State-of-the-Art Comparison}
\label{sec:sota}
Table \ref{tab:sota} presents the performance of different types of models on the standard text-to-video retrieval task, divided into four blocks: Explicit Partial Alignment, Global Matching, Token-wise Matching and our T2VParser based on various baselines. T2VParser demonstrates consistent performance improvements across four benchmark datasets with varying text and video lengths.

The performance gain stems from the framework's ability to utilize powerful pretrained encoders while enabling flexible partial alignment. For short text (e.g., MSR-VTT-1k), the improvements are modest (1.1\%) due to limited multiview embeddings, which rely solely on pretrained encoders. In contrast, complex datasets (e.g., ActivityNet) see larger gains (3.4\%), as multiview information enhances pretrained representations via adaptive partial alignment.

\begin{table*}[t]
\caption{Comparison with Existing Retrieval Methods. The methods use extra tricks (DSL \cite{camoe}) are denoted by the superscript ‘*’ (MSVD do not satisfy the one-to-one retrieval condition). Our Re-implemented methods are denoted by the \dag. The highest retrieval recall in each block is marked with \underline{underline}. The highest overall score is marked with \textbf{bold}. The recall of our models is marked with {\color[HTML]{000AFF} blue} when it is better than the baseline model.}
\vspace{-5pt}
\label{tab:sota}
\resizebox{\textwidth}{!}{%
\begin{tabular}{lcccccccccccccccc}
\hline
\multicolumn{1}{l|}{}                                         & \multicolumn{4}{c|}{\textbf{MSR-VTT 1k}}                                                                                                    & \multicolumn{4}{c|}{\textbf{MSVD}}                                                                                                                                              & \multicolumn{4}{c|}{\textbf{DiDeMo}}                                                                                                        & \multicolumn{4}{c}{\textbf{ActivityNet Caption}}                                                                       \\
\multicolumn{1}{l|}{\multirow{-2}{*}{\textbf{Methods}}}       & \textbf{R@1}                & \textbf{R@5}                & \textbf{R@10}               & \multicolumn{1}{c|}{\textbf{R@sum}}               & \textbf{R@1}                         & \textbf{R@5}                         & \textbf{R@10}                        & \multicolumn{1}{c|}{\textbf{R@sum}}                        & \textbf{R@1}                & \textbf{R@5}                & \textbf{R@10}               & \multicolumn{1}{c|}{\textbf{R@sum}}               & \textbf{R@1}                & \textbf{R@5}                & \textbf{R@10}               & \textbf{R@sum}               \\ \hline
\multicolumn{17}{c}{\textbf{Explicit Partial Alignment}}                                                                                                                                                                                                                                                                                                                                                                                                                                                                                                                                                                                                             \\ \hline
\multicolumn{1}{l|}{HGR (CVPR'20) \cite{HGR}}                             & 9.2                         & 26.2                        & 36.5                        & \multicolumn{1}{c|}{71.9}                         & -                                    & -                                    & -                                    & \multicolumn{1}{c|}{-}                                     & -                           & -                           & -                           & \multicolumn{1}{c|}{-}                            & -                           & -                           & -                           & -                            \\
\multicolumn{1}{l|}{HANet (MM'21) \cite{HANet}}                             & {\ul 9.3}                   & {\ul 27.0}                  & {\ul 38.1}                  & \multicolumn{1}{c|}{{\ul 74.4}}                   & -                                    & -                                    & -                                    & \multicolumn{1}{c|}{-}                                     & -                           & -                           & -                           & \multicolumn{1}{c|}{-}                            & -                           & -                           & -                           & -                            \\ \hline
\multicolumn{17}{c}{\textbf{Global Matching}}                                                                                                                                                                                                                                                                                                                                                                                                                                                                                                                                                                                                                        \\ \hline
\multicolumn{1}{l|}{Frozen (ICCV'21) \cite{frozen_in_time}}                          & 31.0                        & 59.5                        & 70.5                        & \multicolumn{1}{c|}{161.0}                        & 33.7                                 & 64.7                                 & 76.3                                 & \multicolumn{1}{c|}{174.7}                                 & 34.6                        & 65.0                        & 74.7                        & \multicolumn{1}{c|}{174.3}                        & 28.8                        & 60.9                        & -                           & -                            \\
\multicolumn{1}{l|}{CLIP4Clip (Neurocomputing'22) \cite{clip4clip}}             & 44.5                        & 71.4                        & 81.6                        & \multicolumn{1}{c|}{197.5}                        & 45.2                                 & 75.5                                 & 84.3                                 & \multicolumn{1}{c|}{205.0}                                 & 43.4                        & 70.2                        & 80.6                        & \multicolumn{1}{c|}{194.2}                        & 40.5                        & 72.4                        & -                           & -                            \\
\multicolumn{1}{l|}{MPT (MM'24) \cite{MPT}}                               & 49.2                        & 72.9                        & 82.4                        & \multicolumn{1}{c|}{204.5}                        & -                                    & -                                    & -                                    & \multicolumn{1}{c|}{-}                                     & 46.4                        & 72.2                        & 81.4                        & \multicolumn{1}{c|}{200.0}                        & 41.4                        & 70.9                        & 82.9                        & 195.2                        \\
\multicolumn{1}{l|}{STAN* (CVPR'23) \cite{STAN}}                           & {\ul 54.1}                  & {\ul 79.5}                  & {\ul 87.8}                  & \multicolumn{1}{c|}{{\ul 221.4}}                  & -                                    & -                                    & -                                    & \multicolumn{1}{c|}{-}                                     & {\ul 54.6}                  & {\ul 78.4}                  & {\ul 85.1}                  & \multicolumn{1}{c|}{{\ul 218.1}}                  & -                           & -                           & -                           & -                            \\ \hline
\multicolumn{17}{c}{\textbf{Token-wise Matching}}                                                                                                                                                                                                                                                                                                                                                                                                                                                                                                                                                                                                                    \\ \hline
\multicolumn{1}{l|}{Visual Prototype (NeurIPS'22) \cite{VPM}}             & 36.2                        & 64.2                        & 75.7                        & \multicolumn{1}{c|}{176.1}                        & 36.1                                 & 67.4                                 & 81.3                                 & \multicolumn{1}{c|}{184.8}                                 & 36.5                        & 64.9                        & 75.4                        & \multicolumn{1}{c|}{176.8}                        & -                           & -                           & -                           & -                            \\
\multicolumn{1}{l|}{X-CLIP (CVPR'23) \cite{X-CLIP}}                          & 49.3                        & 75.8                        & 84.8                        & \multicolumn{1}{c|}{209.9}                        & 50.4                                 & 80.6                                 & -                                    & \multicolumn{1}{c|}{-}                                     & 47.8                        & 79.3                        & -                           & \multicolumn{1}{c|}{-}                            & 46.2                        & 75.5                        & -                           & -                            \\
\multicolumn{1}{l|}{Cap4Video (CVPR'23) \cite{cap4video}}                       & 51.4                        & 75.7                        & 83.9                        & \multicolumn{1}{c|}{211.0}                        & 51.8                                 & 80.8                                 & 88.3                                 & \multicolumn{1}{c|}{220.9}                                 & 52.0                        & 79.4                        & 87.5                        & \multicolumn{1}{c|}{218.9}                        & -                           & -                           & -                           & -                            \\
\multicolumn{1}{l|}{NarVid (CVPR'25) \cite{narratingthevideo}}                          & 52.7                        & 77.7                        & 85.6                        & \multicolumn{1}{c|}{216.0}                        & 53.1                                 & 81.4                                 & 88.8                                 & \multicolumn{1}{c|}{223.3}                                 & 53.4                        & 79.1                        & 86.3                        & \multicolumn{1}{c|}{218.8}                        & -                           & -                           & -                           & -                            \\
\multicolumn{1}{l|}{Video-ColBERT (CVPR'25) \cite{Video-ColBERT}}                   & 51.5                        & 76.3                        & 85.5                        & \multicolumn{1}{c|}{213.3}                        & 55.2                                 & 82.9                                 & 89.4                                 & \multicolumn{1}{c|}{227.5}                                 & 51.9                        & 78.3                        & 85.6                        & \multicolumn{1}{c|}{215.8}                        & 50.6                        & 78.0                        & 87.9                        & 216.5                        \\
\multicolumn{1}{l|}{CLIP-VIP* (ICLR'23) \cite{clip-vip}}                       & {\ul 57.7}                  & 80.5                        & 88.2                        & \multicolumn{1}{c|}{226.4}                        & -                                    & -                                    & -                                    & \multicolumn{1}{c|}{-}                                     & 55.3                        & 82.0                        & 89.3                        & \multicolumn{1}{c|}{226.6}                        & {\ul 61.4}                  & {\ul 85.7}                  & {\ul 92.6}                  & {\ul 239.7}                  \\
\multicolumn{1}{l|}{Mug-STAN* (arXiv'23) \cite{mug-stan}}                      & 57.3                        & {\ul 81.6}                  & {\ul 88.4}                  & \multicolumn{1}{c|}{{\ul 227.3}}                  & -                                    & -                                    & -                                    & \multicolumn{1}{c|}{-}                                     & {\ul 61.2}                  & {\ul 84.1}                  & {\ul 88.9}                  & \multicolumn{1}{c|}{{\ul 234.2}}                  & -                           & -                           & -                           & -                            \\ \hline
\multicolumn{17}{c}{\textbf{Our Adaptive Partial Alignment}}                                                                                                                                                                                                                                                                                                                                                                                                                                                                                                                                                                                                         \\ \hline
\multicolumn{17}{l}{\cellcolor[HTML]{C0C0C0}\textit{CLIP-ViT-B/32:}}                                                                                                                                                                                                                                                                                                                                                                                                                                                                                                                                                                                                 \\ \hline
\multicolumn{1}{l|}{CLIP4Clip\dag}             & 44.5                        & 71.0                        & 81.6                        & \multicolumn{1}{c|}{197.1}                        & 45.1                                 & 75.5                                 & 83.9                                 & \multicolumn{1}{c|}{204.5}                                 & 43.4                        & 70.1                        & 80.1                        & \multicolumn{1}{c|}{193.6}                        & 40.2                        & 72.4                        & 80.4                        & 193.0                        \\
\multicolumn{1}{l|}{T2VParser+CLIP4Clip\dag}   & {\color[HTML]{000AFF} 46.4} & {\color[HTML]{000AFF} 73.6} & {\color[HTML]{000AFF} 82.4} & \multicolumn{1}{c|}{{\color[HTML]{000AFF} 202.4}} & {\color[HTML]{000AFF} 47.6}          & {\color[HTML]{000AFF} 78.4}          & {\color[HTML]{000AFF} 85.1}          & \multicolumn{1}{c|}{{\color[HTML]{000AFF} 211.1}}          & {\color[HTML]{000AFF} 45.4} & {\color[HTML]{000AFF} 72.4} & {\color[HTML]{000AFF} 83.7} & \multicolumn{1}{c|}{{\color[HTML]{000AFF} 201.5}} & {\color[HTML]{000AFF} 43.7} & {\color[HTML]{000AFF} 74.4} & {\color[HTML]{000AFF} 80.9} & {\color[HTML]{000AFF} 199.0} \\
\multicolumn{1}{l|}{T2VParser+CLIP4Clip\dag *} & 50.2                        & 74.4                        & 84.2                        & \multicolumn{1}{c|}{208.8}                        & -                                    & -                                    & -                                    & \multicolumn{1}{c|}{-}                                     & 49.3                        & 76.4                        & 84.1                        & \multicolumn{1}{c|}{209.8}                        & 47.3                        & 76.5                        & 85.3                        & 209.1                        \\ \hline
\multicolumn{1}{l|}{CLIP-VIP\dag}              & 50.1                        & 74.9                        & 83.1                        & \multicolumn{1}{c|}{208.1}                        & 49.7                                 & 79.7                                 & 85.5                                 & \multicolumn{1}{c|}{214.9}                                 & 48.3                        & 77.0                        & 84.4                        & \multicolumn{1}{c|}{209.7}                        & 51.1                        & 78.6                        & 87.9                        & 217.6                        \\
\multicolumn{1}{l|}{T2VParser+CLIP-VIP\dag}    & {\color[HTML]{000AFF} 51.3} & {\color[HTML]{000AFF} 74.8} & {\color[HTML]{000AFF} 84.3} & \multicolumn{1}{c|}{{\color[HTML]{000AFF} 210.4}} & {\color[HTML]{000AFF} 52.8}          & {\color[HTML]{000AFF} 81.3}          & {\color[HTML]{000AFF} 86.3}          & \multicolumn{1}{c|}{{\color[HTML]{000AFF} 220.4}}          & {\color[HTML]{000AFF} 52.3} & {\color[HTML]{000AFF} 77.4} & {\color[HTML]{000AFF} 85.7} & \multicolumn{1}{c|}{{\color[HTML]{000AFF} 215.4}} & {\color[HTML]{000AFF} 55.6} & {\color[HTML]{000AFF} 82.4} & {\color[HTML]{000AFF} 89.4} & {\color[HTML]{000AFF} 227.4} \\
\multicolumn{1}{l|}{T2VParser+CLIP-VIP\dag *}  & 55.2                        & 78.1                        & 87.4                        & \multicolumn{1}{c|}{220.7}                        & -                                    & -                                    & -                                    & \multicolumn{1}{c|}{-}                                     & 55.6                        & 79.9                        & 87.1                        & \multicolumn{1}{c|}{222.6}                        & 60.2                        & 84.7                        & 91.9                        & 236.8                        \\ \hline
\multicolumn{1}{l|}{Mug-STAN\dag}              & 50.9                        & 74.4                        & 84.3                        & \multicolumn{1}{c|}{209.6}                        & 51.9                                 & 80.1                                 & 85.4                                 & \multicolumn{1}{c|}{217.4}                                 & 52.4                        & 77.9                        & 85.3                        & \multicolumn{1}{c|}{215.6}                        & 48.7                        & 73.6                        & 84.0                        & 206.3                        \\
\multicolumn{1}{l|}{T2VParser+Mug-STAN\dag}    & {\color[HTML]{000AFF} 51.6} & {\color[HTML]{000AFF} 74.9} & {\color[HTML]{000AFF} 84.7} & \multicolumn{1}{c|}{{\color[HTML]{000AFF} 211.2}} & {\color[HTML]{000AFF} 53.1}          & {\color[HTML]{000AFF} 82.5}          & {\color[HTML]{000AFF} 86.7}          & \multicolumn{1}{c|}{{\color[HTML]{000AFF} 222.3}}          & {\color[HTML]{000AFF} 53.2} & {\color[HTML]{000AFF} 78.1} & {\color[HTML]{000AFF} 86.2} & \multicolumn{1}{c|}{{\color[HTML]{000AFF} 217.5}} & {\color[HTML]{000AFF} 51.0} & {\color[HTML]{000AFF} 76.2} & {\color[HTML]{000AFF} 84.5} & {\color[HTML]{000AFF} 211.7} \\
\multicolumn{1}{l|}{T2VParser+Mug-STAN\dag *}  & 55.7                        & 78.4                        & 88.1                        & \multicolumn{1}{c|}{222.2}                        & -                                    & -                                    & -                                    & \multicolumn{1}{c|}{-}                                     & 58.7                        & 80.3                        & 87.6                        & \multicolumn{1}{c|}{226.6}                        & 55.3                        & 76.7                        & 87.1                        & 219.1                        \\ \hline
\multicolumn{17}{l}{\cellcolor[HTML]{C0C0C0}\textit{CLIP-ViT-B/16:}}                                                                                                                                                                                                                                                                                                                                                                                                                                                                                                                                                                                                 \\ \hline
\multicolumn{1}{l|}{CLIP-VIP\dag}              & 54.1                        & 77.0                        & 84.7                        & \multicolumn{1}{c|}{215.8}                        & 52.2                                 & 81.6                                 & 88.2                                 & \multicolumn{1}{c|}{222.0}                                 & 50.5                        & 78.3                        & 86.6                        & \multicolumn{1}{c|}{215.4}                        & 53.4                        & 82.3                        & 89.7                        & 225.4                        \\
\multicolumn{1}{l|}{T2VParser+CLIP-VIP\dag}    & {\color[HTML]{000AFF} 55.2} & {\color[HTML]{000AFF} 78.6} & {\color[HTML]{000AFF} 84.9} & \multicolumn{1}{c|}{{\color[HTML]{000AFF} 218.7}} & {\color[HTML]{000AFF} 54.0}          & {\color[HTML]{000AFF} \textbf{83.7}} & {\color[HTML]{000AFF} \textbf{88.8}} & \multicolumn{1}{c|}{{\color[HTML]{000AFF} 226.5}}          & {\color[HTML]{000AFF} 52.3} & {\color[HTML]{000AFF} 79.8} & {\color[HTML]{000AFF} 87.4} & \multicolumn{1}{c|}{{\color[HTML]{000AFF} 219.5}} & {\color[HTML]{000AFF} 56.8} & {\color[HTML]{000AFF} 84.3} & {\color[HTML]{000AFF} 91.7} & {\color[HTML]{000AFF} 232.8} \\
\multicolumn{1}{l|}{T2VParser+CLIP-VIP\dag *}  & {\ul \textbf{58.4}}         & {\ul \textbf{82.1}}         & 88.1                        & \multicolumn{1}{c|}{228.6}                        & -                                    & -                                    & -                                    & \multicolumn{1}{c|}{-}                                     & 57.0                        & 85.1                        & 89.6                        & \multicolumn{1}{c|}{231.7}                        & {\ul \textbf{61.8}}         & 86.7                        & 92.3                        & 240.8                        \\ \hline
\multicolumn{1}{l|}{Mug-STAN\dag}              & 53.4                        & 76.4                        & 85.2                        & \multicolumn{1}{c|}{215.0}                        & 53.3                                 & 82.4                                 & 87.0                                 & \multicolumn{1}{c|}{222.7}                                 & 55.7                        & 78.6                        & 87.3                        & \multicolumn{1}{c|}{221.6}                        & 51.2                        & 82.4                        & 88.4                        & 222.0                        \\
\multicolumn{1}{l|}{T2VParser+Mug-STAN\dag}    & {\color[HTML]{000AFF} 54.5} & {\color[HTML]{000AFF} 78.4} & {\color[HTML]{000AFF} 85.6} & \multicolumn{1}{c|}{{\color[HTML]{000AFF} 218.5}} & {\color[HTML]{000AFF} \textbf{54.3}} & {\color[HTML]{000AFF} \textbf{83.7}} & {\color[HTML]{000AFF} 88.7}          & \multicolumn{1}{c|}{{\color[HTML]{000AFF} \textbf{226.7}}} & {\color[HTML]{000AFF} 56.9} & {\color[HTML]{000AFF} 80.3} & {\color[HTML]{000AFF} 86.8} & \multicolumn{1}{c|}{{\color[HTML]{000AFF} 224.0}} & {\color[HTML]{000AFF} 54.4} & {\color[HTML]{000AFF} 84.1} & {\color[HTML]{000AFF} 90.7} & {\color[HTML]{000AFF} 229.2} \\
\multicolumn{1}{l|}{T2VParser+Mug-STAN\dag *}  & {\ul \textbf{58.4}}         & {\ul \textbf{82.1}}         & {\ul \textbf{88.4}}         & \multicolumn{1}{c|}{{\ul \textbf{228.9}}}         & -                                    & -                                    & -                                    & \multicolumn{1}{c|}{-}                                     & {\ul \textbf{62.4}}         & {\ul \textbf{85.2}}         & {\ul \textbf{89.5}}         & \multicolumn{1}{c|}{{\ul \textbf{237.1}}}         & 60.5                        & {\ul \textbf{89.7}}         & {\ul \textbf{93.5}}         & {\ul \textbf{243.7}}         \\ \hline
\end{tabular}
}
\end{table*}

\subsection{Partial Alignment under Noise Content}
\begin{table}[]
\caption{Evaluation of the partial alignment with noise content on ActivityNet and PRVR benchmark (TV show retrieval dataset). With the CLIP-ViT-B/32 as backbone.}
\vspace{-5pt}
\label{tab:nc}
\begin{tabular}{l|cc}
\hline
\textbf{Method}     & \textbf{ActivityNet-Noise} & \textbf{PRVR} \\ \hline
CLIP4Clip           & 21.3                       & 19.7                              \\
T2VParser+CLIP4Clip & 32.5                       & 33.4                              \\
Improvement         & \textbf{+11.2}             & \textbf{+13.7}                    \\ \hline
CLIP-VIP            & 41.1                       & 33.4                              \\
T2VParser+CLIP-VIP  & 45.2                       & 40.5                              \\
Improvement         & \textbf{+4.1}              & \textbf{+7.1}                     \\ \hline
Mug-STAN            & 38.5                       & 32.6                              \\
T2VParser+Mug-STAN  & 42.1                       & 39.1                              \\
Improvement         & \textbf{+3.6}              & \textbf{+6.5}                     \\ \hline
\end{tabular}
\vspace{-5pt}
\end{table}

T2VParser effectively reduces noise from content inconsistency and partial alignment. Sec \ref{4.1} and \ref{sec:sota} covers aligned text-video cases, while mismatches (where either modality fails to fully reflect the other) are common in real-world data (e.g., YouTube metadata). To simulate textual noise, we modified the query composition in the ActivityNet Captions dataset. For each video $v$, we randomly replaced 25\% of its original captions $\{ c_1, c_2, ..., c_n \}$ with captions $c'$ from other videos, constructing new queries $t'=concat(c_1, c_2', ..., c_k', ...,c_n)$. Besides, we also tested visual noise where videos contain rapidly changing scenes, extended durations, and only partial relevance to the textual descriptions. For evaluation, we adopted the benchmark from Partially Relevant Video Retrieval (PRVR) \cite{PRVR}. As shown in Table \ref{tab:nc}, T2VParser demonstrates improved robustness across different baselines. This improvement stems from (1) identifying correct visual-textual alignment during training and (2) filtering out noisy content through partial alignment during matching.


\subsection{Analysis of Multiview Embeddings}
\begin{figure}[t]
  \centering
  \includegraphics[width=\linewidth]{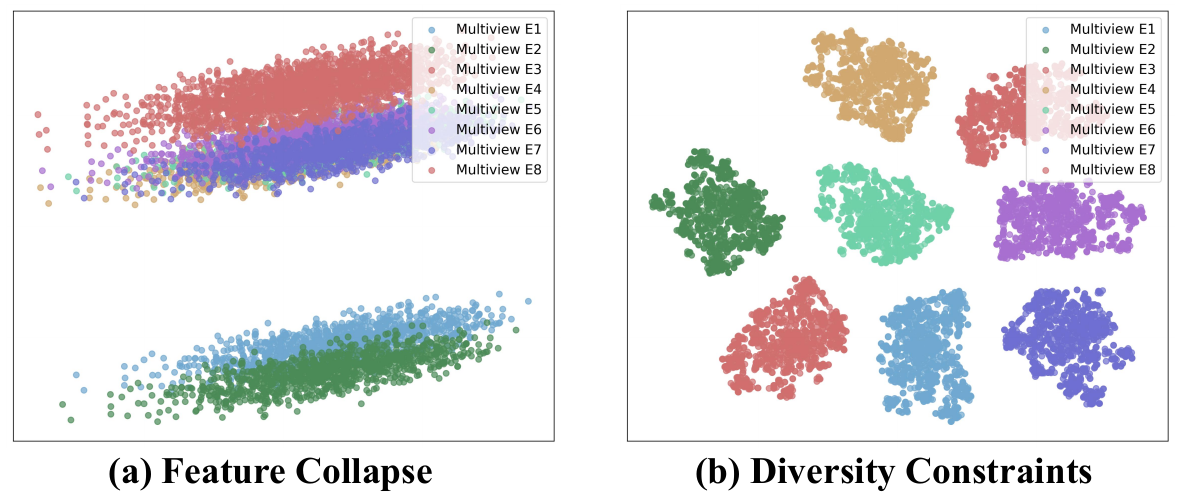}
  \vspace{-15pt}
  \caption{Analysis of Multiview Embeddings. Each color represents embeddings from a different adaptive decomposition token. Each point within a group represents a sample.}
  \label{fig:vis1}
  \vspace{-15pt}
\end{figure}

This section analyzes Adaptive Decomposition Tokens (ADTs) and multiview embeddings in T2VParser. While pretrained encoders provide foundational alignment, ADTs decompose multiview semantics through  constraints. Without $L_{div}$, Figure \ref{fig:vis1}(a) shows feature collapse causing redundant representations. With $L_{div}$ (Figure \ref{fig:vis1}(b)), embeddings capture diverse details, enabling effective partial alignment via the Dual Communication Mechanism.

\begin{figure}[]
  \centering
  \includegraphics[width=\linewidth]{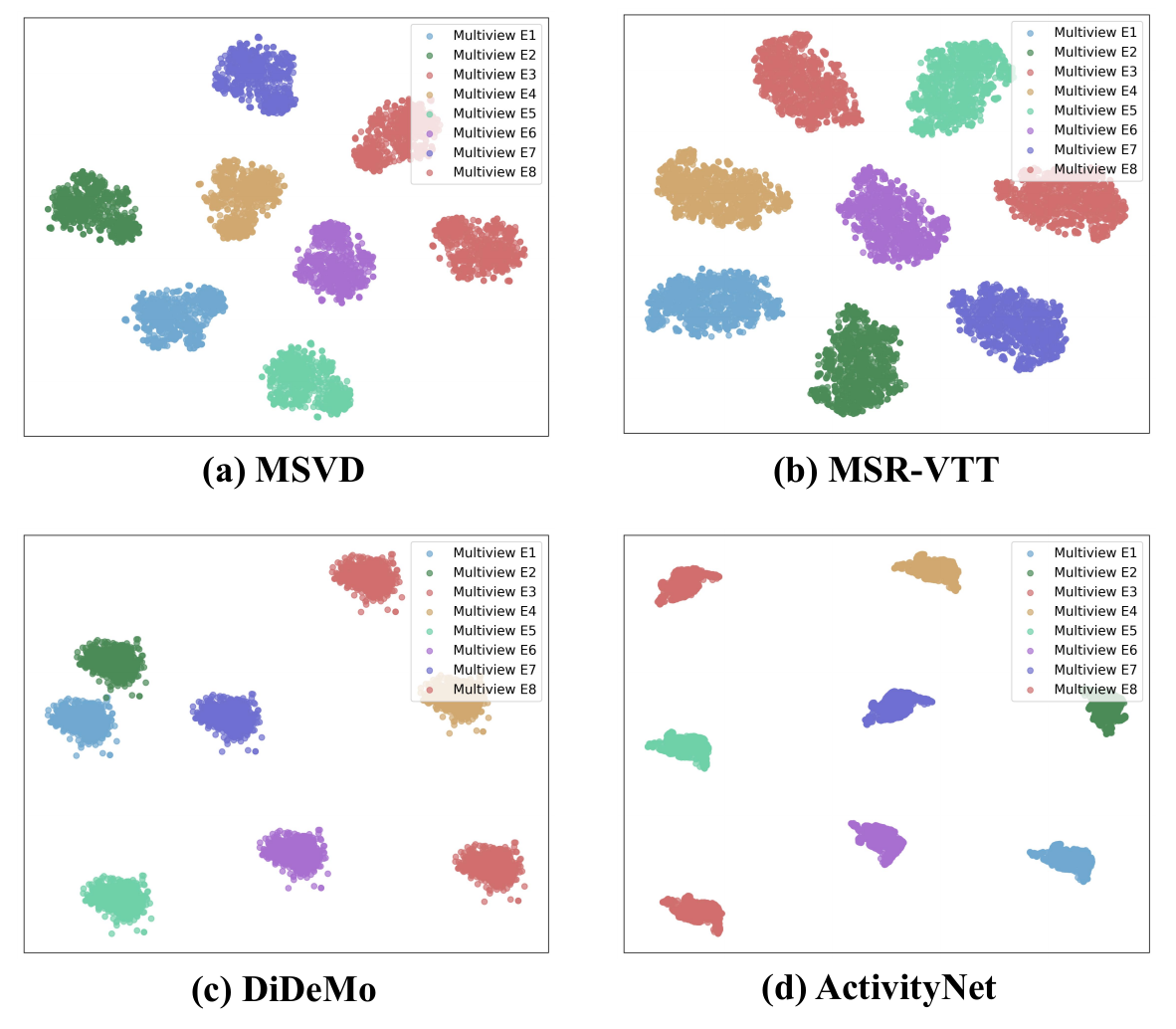}
  \vspace{-15pt}
  \caption{Impact of Modality Content on Multiview Embeddings. From (a) to (d), the video length gradually increases. }
  \label{fig:vis2}
  \vspace{-10pt}
\end{figure}

Sec \ref{4.1} and \ref{sec:sota} highlights how modality content richness affects multiview embedding extraction. Table \ref{tab:differentvl} demonstrates that longer videos yield more diverse representations (the similarity between each multiview embedding drops from 0.29 to 0.04), reducing feature collapse (Figure \ref{fig:vis2}). This explains T2VParser's limited gains on data-scarce MSR-VTT 1k (Table \ref{tab:sd}), as ADTs require sufficient information complexity to effectively disentangle local details.

\begin{table}[]
\caption{Evaluation of Multiview Embeddings with Different Video Length.}
\vspace{-5pt}
\label{tab:differentvl}
\begin{tabular}{l|cc}
\hline
\textbf{Dataset} & \textbf{Video Length(s)↑} & \textbf{Multiview Similarity↓} \\ \hline
MSVD             & 9.65                     & 0.29                           \\
MSR-VTT 1k       & 14.86                      & 0.27                           \\
DiDeMo           & 54.10                     & 0.08                           \\
ActivityNet      & 116.55                    & 0.04                           \\ \hline
\end{tabular}
\end{table}

\subsection{Ablation Study on Model Structure}
\label{ablms}
\begin{table}[]
\caption{Ablation Study on MSR-VTT 1k. Re-implemented methods are
denoted by the \dag.}
\vspace{-5pt}
\label{tab:abl}
\begin{tabular}{llll}
\hline
\multicolumn{1}{l|}{\textbf{Method}}                    & \textbf{R1}   & \textbf{R5}   & \textbf{R10}  \\ \hline
\multicolumn{1}{l|}{\textbf{T2VParser(full model)}}     & \textbf{58.4} & \textbf{82.1} & \textbf{88.4} \\ \hline
\multicolumn{4}{c}{\textit{w/o Dual Communication Mechanism}}                                           \\ \hline
\multicolumn{1}{l|}{w/ Max Pooling}                     & 52.4          & 76.5          & 82.6          \\
\multicolumn{1}{l|}{w/ Mean Pooling}                    & 53.1          & 77.1          & 84.2          \\ \hline
\multicolumn{4}{c}{\textit{w/o Multiview Embeddings}}                                                   \\ \hline
\multicolumn{1}{l|}{w/ Global Embedding}                & 55.2          & 79.7          & 87.0          \\
\multicolumn{1}{l|}{w/ Local Embeddings}                & 57.0          & 81.7          & 88.2          \\
\multicolumn{1}{l|}{w/ Global and Local Embeddings}     & 56.8          & 80.9          & 87.6          \\ \hline
\multicolumn{4}{c}{\textit{Ablation on Doc-Video Training}}                                             \\ \hline
\multicolumn{1}{l|}{w/o Document Generation}            & 57.6          & 81.6          & 88.1          \\
\multicolumn{1}{l|}{Mug-STAN\dag (baseline)}               & 57.0          & 81.2          & 87.9          \\
\multicolumn{1}{l|}{w/ Document Generation in baseline} & 57.2          & 81.3          & 88.0          \\ \hline
\multicolumn{4}{c}{\textit{Others}}                                                                     \\ \hline
\multicolumn{1}{l|}{w/o Representation Diversity Loss}  & 56.8          & 79.6          & 87.4          \\
\multicolumn{1}{l|}{w/o Local Representation}           & 54.7          & 78.3          & 87.0          \\ \hline
\end{tabular}
\vspace{-5pt}
\end{table}

As demonstrated in Table \ref{tab:abl}, four different experimental configurations were designed to evaluate various model settings.

\noindent \textbf{w/o Dual Communication Mechanism.} 
Experimental results confirm the Dual Communication Mechanism's effectiveness in complementary information extraction and redundancy removal. Simplified pooling approaches (w/ Max Pooling: 52.4\%, w/ Mean Pooling: 53.1\%) suffer from information loss and lack cross representation communication, despite computational efficiency.

\noindent \textbf{w/o Multiview Embeddings.}
This block evaluates the role of multiview embedding in alignment through ablation studies. Three configurations were tested: (1) w/ Global Embedding utilizes only global embeddings from the pretrained encoder with cosine similarity for alignment, resembling CLIP4clip's approach \cite{clip4clip} but susceptible to data partial misalignment; (2) w/ Local Embedding employs token/frame embeddings processed through the Dual Communication Mechanism; (3) w/ Global and Local Embeddings concatenates both embedding types before Dual Communication processing. While the latter two configurations, comparable to CLIP-VIP's performance, address detailed information and partial alignment, they treat tokens and frames as independent units, neglecting holistic video information. This limitation results in performance degradation from 58.4\% to 57.0\%, as discussed in Section \ref{sec:intro}.

\noindent \textbf{Ablation on Doc-Video Training.}
The Doc-Video Training enhances T2VParser's multiview embedding decomposition, with experiments showing its performance gains stem not merely from extra descriptions but from effective multiview embedding integration. Testing without document generation (w/o Document Generation) yielded a 0.8\% performance drop in T2VParser. While Mug-STAN, a sota token-wise interaction model, improved only marginally (57.0\% to 57.2\%) when trained with the same Doc-Video strategy, and T2VParser also beyond baseline. These results confirm that (1) T2VParser's advantage lies in its multiview embeddings' complementary retrieval effects, and (2) merely adding document descriptions fails to substantially boost performance across architectures.

\noindent \textbf{Others.}
Ablation studies of other key configurations revealed: w/o Representation Diversity Loss, multiview embeddings suffered feature collapse, impairing alignment. w/o Local Representation underutilized the encoder's local features and lost detailed information by relying solely on multiview embeddings.

\subsection{Ablation on Different LLMs}
Document generation enhances model generalization and provides richer training data for text parser optimization. Experiments reveal consistent performance across LLMs (Deepseek-v2 \cite{deepseek}, LLaMA \cite{llama}, ChatGLM \cite{chatglm}), with <0.3\% accuracy variation on MSR-VTT (Table \ref{tab:ablm}), demonstrating architecture-agnostic effectiveness.

\begin{table}[]
\caption{Ablation Study on different LLMs. We use T2VParser+Mug-STAN encoder with MSR-VTT 1k benchmark. Each LLM selects the latest version}
\label{tab:ablm}
\vspace{-5pt}
\centering
\begin{tabular}{l|ccc}
\hline
\textbf{LLM} & \textbf{R@1} & \textbf{R@5} & \textbf{R@10} \\ \hline
Deepseek-v2 \cite{deepseek}  & 58.4         & 82.1         & 88.4          \\
LLaMA \cite{llama}       & 58.1         & 82.0         & 88.3          \\
ChatGLM \cite{chatglm}     & 58.4         & 81.9         & 88.4          \\ \hline
\end{tabular}
\vspace{-5pt}
\end{table}

\subsection{Parameter Analysis}
\begin{figure}[t]
  \centering
  \includegraphics[width=\linewidth]{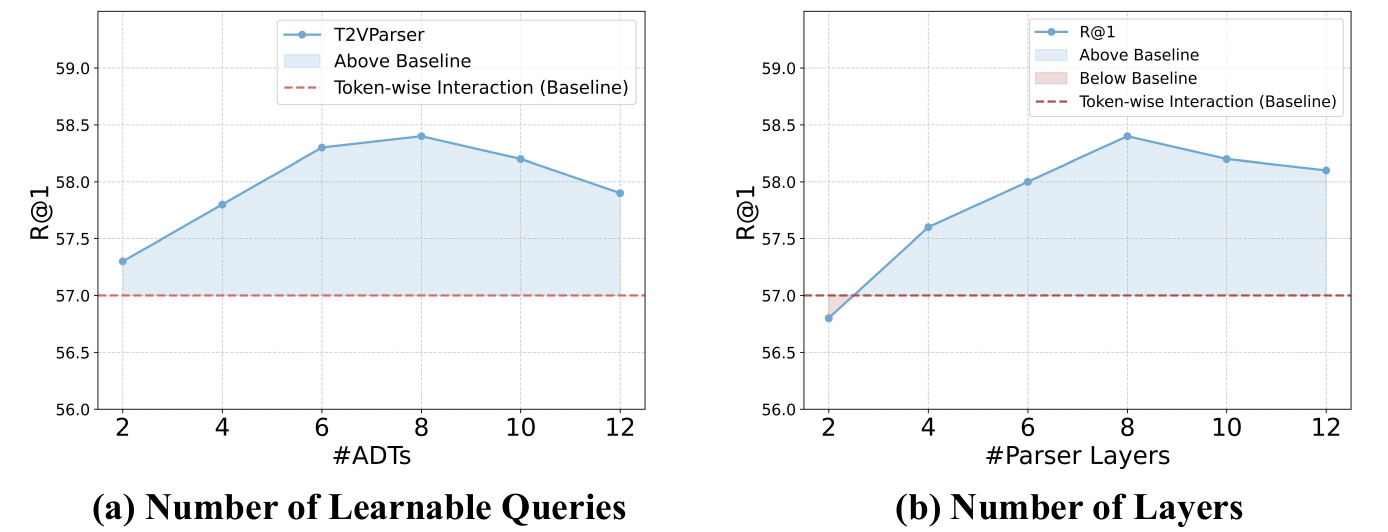}
  \vspace{-15pt}
  \caption{Different Number of Learnable Queries and Layers of T2VParser(w/ Mug-STAN encoders) on MSR-VTT 1k.}
  \label{fig:para}
  \vspace{-10pt}
\end{figure}

The evaluation of adaptive decomposition tokens (ADTs) quantity and parser depth (Figure \ref{fig:para}) demonstrates that our model consistently outperforms the baseline (red), achieving peak recall (58.4\%) with 8 tokens (an optimal balance between information redundancy and representation capacity) while shallow architectures (2 layers) underperform by 0.2\% due to insufficient semantic preservation, and deeper 8-layer configurations attain maximum performance through comprehensive feature extraction and integration.

\section{Limitation and Future Work}

T2VParser's superior decomposition and alignment performance comes with computational trade-offs: while effectively leveraging pretrained encoders, its additional parser and Dual Communication Mechanism introduce 15\% longer per-query retrieval times versus CLIP-B16 baselines, with efficiency gaps scaling proportionally to candidate set size (Table \ref{tab:lim}). Evaluations across MSR-VTT (1K/3K videos) confirm this accuracy-efficiency balance.
While multiview embeddings capture independent semantics, their implicit nature limits interpretability and flexibility. Future work will: (1) optimize retrieval efficiency via generative models (e.g., T2VIndexer \cite{t2vindexer}) for cost-effective fine-grained search, and (2) improve embedding interpretability for broader application scenarios.

\begin{table}[]
\caption{Retrieval Efficiency Comparative Analysis on MSR-VTT. Re-implemented methods are
denoted by the \dag. {\color[HTML]{000AFF} Blue} means Text/Video Parser time cost, and {\color[HTML]{FE0000} red} means Dual Communication Mechanism time cost.}
\label{tab:lim}
\vspace{-5pt}
\begin{tabular}{l|cc|cc}
\hline
\multicolumn{1}{c|}{\multirow{2}{*}{\textbf{Method}}} & \multicolumn{2}{c|}{\textbf{\#Candidate 1000}}                                                              & \multicolumn{2}{c}{\textbf{\#Candidate 3000}}                                                               \\ \cline{2-5} 
\multicolumn{1}{c|}{}                                 & \multicolumn{1}{c|}{\textbf{\begin{tabular}[c]{@{}c@{}}Inference\\ Time(ms)↓\end{tabular}}} & \textbf{R@1↑} & \multicolumn{1}{c|}{\textbf{\begin{tabular}[c]{@{}c@{}}Inference\\ Time(ms)↓\end{tabular}}} & \textbf{R@1↑} \\ \hline
CLIP-VIP\dag                                             & 192                                                                                         & 57.3          & 584                                                                                         & 51.3          \\
STAN\dag                                                 & 194                                                                                         & 54.6          & 603                                                                                         & 46.9          \\
Mug-STAN\dag                                             & 246                                                                                         & 57.0          & 735                                                                                         & 50.6          \\ \hline
\textbf{T2VParser}                                    & \textbf{283 ({\color[HTML]{000AFF} 21}+{\color[HTML]{FE0000} 26})}                                                                         & \textbf{58.4} & \textbf{834 ({\color[HTML]{000AFF} 62}+{\color[HTML]{FE0000} 81})}                                                                         & \textbf{51.9} \\ \hline
\end{tabular}
\vspace{-10pt}
\end{table}

\section{Conclusion}

In this paper, the inherent limitations of video-text data that hinder semantic alignment model training, particularly the partial alignment problem, were systematically analyzed. To address this challenge, T2VParser was proposed, a framework that achieves adaptive semantic decomposition and partial alignment through modality-shared learnable tokens. The representation capabilities of pretrained encoders were preserved to leverage the power of large-scale multimodal alignment models. Furthermore, a Dual Communication Mechanism was designed to facilitate information exchange and redundancy elimination among independent multiview embeddings.
While T2V Parser demos effectiveness based on different encoder structures and data formats, its post-processing approach impacts retrieval efficiency. Additionally, the decomposed multiview embeddings lack explicit representation of partial information. Future work will focus on two key aspects: (1) enhancing interaction efficiency, and (2) improving the interpretability of multiview embeddings to enable individual embedding utilization, thereby increasing system flexibility.

\begin{acks}
This work was supported by the Climbing Plan Project on Information Disclosure Review Technology (Grant No.E5Z0051106).
\end{acks}

\bibliographystyle{ACM-Reference-Format}
\balance
\bibliography{sample-base}










\end{document}